\documentclass[manuscript,acmlarge,nonacm]{acmart}

\usepackage{multirow}
\usepackage{booktabs} 
\usepackage{array} 
\usepackage{paralist} 
\usepackage{verbatim} 
\usepackage{lipsum}  
\usepackage{amsmath}
\usepackage{amsfonts}
\usepackage{url}

\renewcommand\lvec[1]{\overrightarrow{\mathbf{#1}}}
\newcommand\rvec[1]{\overleftarrow{\mathbf{#1}}}
\newcommand*\concat{\hspace{0.2em}\mathbin{\|}\hspace{0.2em}}

\renewcommand\vec[1]{\mathbf{#1}}

\newcommand\matrow[3][]{{#2}{[#3]}^{#1}}

\newcommand\RR{\mathbb{R}}
\newcommand{\nth}[1]{${#1}^\text{th}$}

\newcommand\set[1]{\mathcal{#1}}

\usepackage{tikz,pgfplots}

\newcommand\LSTMR{\overrightarrow{\text{LSTM}}}
\newcommand\LSTML{\overleftarrow{\text{LSTM}}}
\newcommand\ASO{\tilde{A}_{\text{Sub} \leftrightarrow \text{Obj}}}

\AtBeginDocument{%
  \providecommand\BibTeX{{%
    \normalfont B\kern-0.5em{\scshape i\kern-0.25em b}\kern-0.8em\TeX}}}

\acmJournal{HEALTH}

\begin{document}

\title{Attention-Gated Graph Convolutions for Extracting Drug Interaction Information from Drug Labels}

\author{Tung Tran}
\email{tung.tran@uky.edu}
\affiliation{%
  \institution{University of Kentucky}
  \department{Department of Computer Science}
  \city{Lexington}
  \state{Kentucky}
  \country{United States}
}

\author{Ramakanth Kavuluru}
\email{ramakanth.kavuluru@uky.edu}
\affiliation{%
  \institution{University of Kentucky}
  \department{Department of Internal Medicine, Division of Biomedical Informatics}
  \city{Lexington}
  \state{Kentucky}
  \country{United States}
}

\author{Halil Kilicoglu}
\email{halil.kilicoglu@nih.gov}
\affiliation{%
  \institution{National Library of Medicine}
  \department{Lister Hill National Center for Biomedical Communications}
  \city{Bethesda}
  \state{Maryland}
  \country{United States}
}

\renewcommand{\shortauthors}{Tran et al.}

\begin{abstract}
Preventable adverse events as a result of medical errors present a growing concern in the healthcare system. As drug-drug interactions (DDIs) may lead to preventable adverse events, being able to extract DDIs from drug labels into a machine-processable form is an important step toward effective dissemination of drug safety information. Herein, we tackle the problem of jointly extracting drugs and their interactions, including interaction \emph{outcome}, from drug labels. Our deep learning approach entails composing various intermediate representations, including graph-based context derived using graph convolutions (GC) with a novel attention-based gating mechanism (holistically called GCA), which are combined in meaningful ways to predict on all subtasks jointly. Our  model is trained and evaluated on the 2018 TAC DDI corpus. Our GCA model in conjunction with transfer learning performs at 39.20\% F1 and 26.09\% F1 on entity recognition (ER) and relation extraction (RE) respectively on the first official test set and at 45.30\% F1 and 27.87\% F1 on ER and RE respectively on the second official test set corresponding to an improvement over our prior best results by up to 6 absolute F1 points. After controlling for available training data, the proposed model exhibits state-of-the-art performance for this task.
\end{abstract}


\maketitle

\section{Introduction}\label{sec-intro}

Preventable adverse events (AE) are negative consequences of medical care resulting in injury or illness in a way that is generally considered avoidable. According to a report~\citep{levinson2010adverse} by the Department of Human and Health Services, based on an analysis of hospital visits by over a million Medicare beneficiaries, about \emph{one in seven} hospital visits were associated with an AE with 44\% being considered clearly or likely preventable. Overall, AEs were responsible for an estimated US \$324 million in Medicare spending for the studied month of October 2008. Preventable AEs thus introduce a growing concern in the modern healthcare system as they represent a significant fraction of hospital admissions and play a significant role in increased health care costs. Alarmingly, preventable AEs have been cited as the eighth leading cause of death in the U.S., with an estimated fatality rate of between 44,000 and 98,000 each year~\citep{kohn2000err}. As drug-drug interactions (DDIs) may lead to to variety of preventable AEs, being able to extract DDIs from prescription drug labels is an important effort toward effective dissemination of drug safety information. This includes extracting information such as adverse drug reactions and drug-drug interactions as indicated by drug labels. The U.S. Food and Drug Administration (FDA), for example, has recently begun to transform Structured Product Labeling (SPL) documents into a computer-readable format, encoded in national standard terminologies, that will be made available to the the medical community and the public~\citep{fushman2018overview}. The initiative to develop a database of structured drug safety information that can be indexed, searched, and sorted is an important milestone toward a \emph{fully-automated} health information exchange system. 

\begin{figure}[!tpb]
  \centering
  \includegraphics[width=0.65\columnwidth]{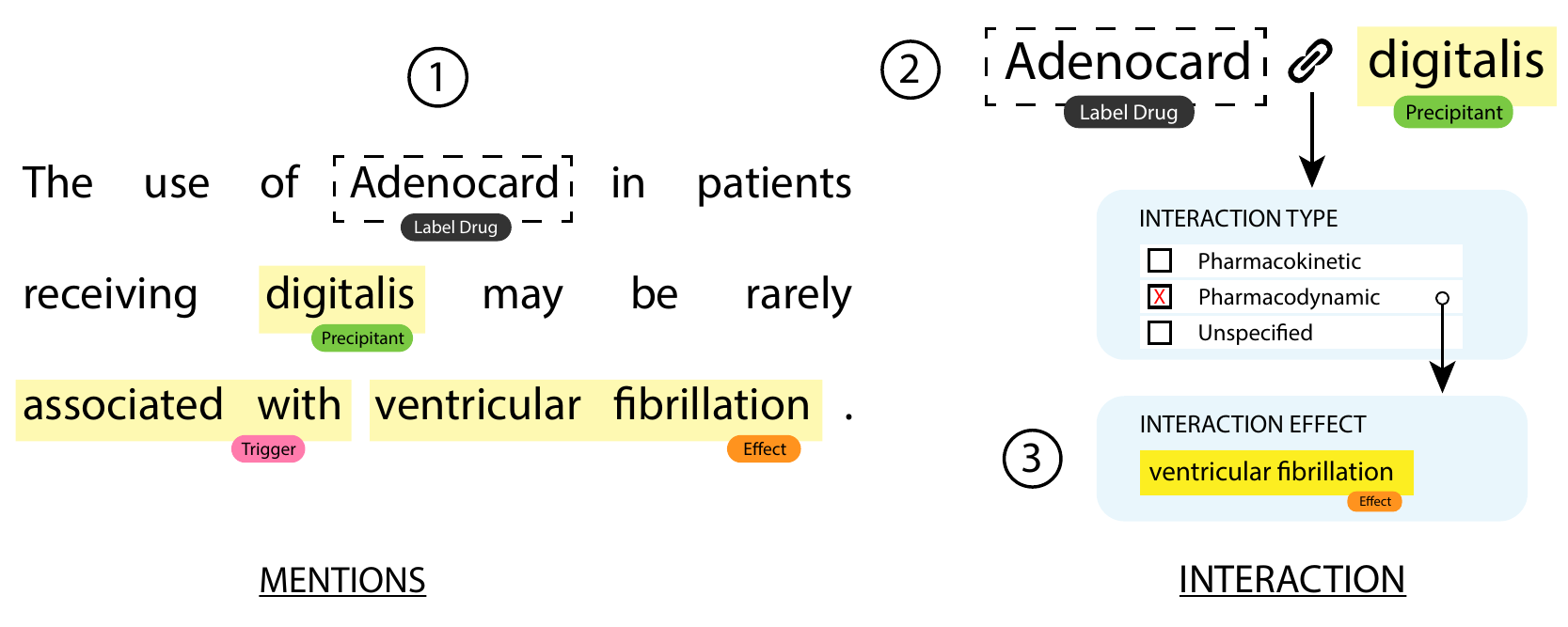}
  \caption{An example illustrating the end-to-end DDI extraction task. We first (1) identify mentions including precipitants; for each precipitant, we (2) determine the type of interaction and, based on interaction type, (3) determine the interaction outcome. In the case of PD interactions, the outcome corresponds to one of the previously identified \emph{effect} spans.}
  \label{fig_tac_example}
  \end{figure}

To aid in this effort, we propose a supervised deep learning model able to tackle the problem of drug-drug interaction extraction in an \emph{end-to-end} fashion. While most prior efforts assume all drug entities are known ahead of time (more in Section~\ref{sec-related}), and the drug-drug interaction extraction task reduces to a simpler \emph{binary relation classification} task of known drug pairs, we propose a system able to identify drug mentions in addition to their interactions. Concretely, the system takes as input the textual content of the label (indicating dosage and drug safety precautions) of a target drug and, as output, identifies mentions of other drugs that interact with the target drug. Thus only one of the two interacting drugs is known beforehand (i.e., the ``label drug''), while the other (i.e., the ``precipitating drug'', or simply precipitant) is an unknown that our model is expected to extract. Along with identifying precipitants, we also determine the type of interaction associated with each precipitant; that is, whether the interaction is designated as being pharmacodynamic (PD) or pharmacokinetic (PK). In pharmacology, PD interactions are associated with a consequence on the organism while PK interactions are associated with changes in how one or both of the interacting drugs is absorbed, transported, distributed, metabolized, and excreted when used jointly. Beyond identifying the interaction type, it is also important to identify the outcome or consequence of an interaction. As defined, PK consequence can be captured using a small fixed vocabulary, while identifying PD effects is a much more contrived process. The latter involves additionally identifying spans of text correspond to a mention of a PD effect and linking each identified PD precipitants to one or more PD effects. We provide a more formal description of the task in Section~\ref{sec-task-description}. Figure~\ref{fig_tac_example} features a simple example of a PD interaction that is extracted from the drug label for \textbf{Adenocard}, where the precipitant is \emph{digitalis} and the effect is ``ventricular fibrillation.''

To address this end-to-end variant of DDI extraction, we propose a multi-task joint-learning architecture wherein various intermediate hidden representations, including sequence-based and graph-based contextual representations based on bidirectional Long Short-Term Memory (BiLSTM) networks and graph convolution (GC) networks respectively, are composed and are then combined in clever ways to produce predictions for each subtask. GCs over dependency parse trees are useful for capturing long-distance syntactic dependencies. We innovate on conventional GCs with a sigmoid gating mechanism derived via additive attention, referred to as Graph Convolution with Attention-Gating (GCA), which determines whether or not (and to what extent) information propagates between source and target nodes corresponding to edges in the dependency tree. The attention component controls information flow by producing a sigmoid gate (corresponding to a value in $[0,1]$) for each edge based on an attention-like mechanism that measures relevance between node pairs. Intuitively, some dependency edges are more relevant than others; for example, \emph{negations} or \emph{adjectives} linked to important nouns via dependency edges may have a large influence on the overall meaning of a sentence while \emph{articles}, such as ``the'', ``a'', and ``an'', have little or no influence comparatively. A standard GC would compose all source nodes with equal weighting, while the GCA would be more selective by possibly assigning a higher sigmoid value to \emph{negations}/\emph{adjectives} and a lower sigmoid value to \emph{articles}. 

\begin{table*}[!t]
	\caption{Characteristics of various datasets\label{tb_datastats}}
	\renewcommand{\arraystretch}{1}
  \resizebox{\textwidth}{!}{
  \begin{tabular}{@{\extracolsep{1em}}l rrrrr}
  \toprule
  & \textbf{*}DDI2013 & \textbf{*}NLM180 & TR22 & Test Set 1 & Test Set 2\\
  \midrule
  Number of Drug Labels & 715 & 180 & 22 & 57 & 66\\
  Total number of sentences & 6489 & 5757 & 603 & 8195 & 4256\\
  Number of sentences per Drug Label (Average) & 9 & 32 & 27 & 144 & 64\\
  Number of words per sentence (Average) & 21 & 23 & 24 & 22 & 23\\
  Proportion of sentences with \emph{annotations} & 70\% & 27\% & 51\% & 23\% & 23\%\\
  Number of mentions per \emph{annotated} sentence (Average) & 2.3 & 4.0 & 3.8 & 3.7 & 3.6\\
  \midrule
  Proportion of mentions that are Precipitant & 100\% & 57\% & 53\% & 56\% & 55\%\\
  Proportion of mentions that are Trigger & - & 20\% & 28\% & 30\% & 33\%\\
  Proportion of mentions that are Effect & - & 23\% & 19\% & 14\% & 12\%\\
  \midrule
  Proportion of interactions that are Pharmacodynamic  & 14\% & 47\% & 49\% & 33\% & 28\%\\
  Proportion of interactions that are Pharmacokinetic & 9\% & 25\% & 21\% & 28\% & 47\%\\
  Proportion of interactions that are Unspecified  & 77\% & 28\% & 30\% & 39\% & 25\%\\
  \bottomrule
  \end{tabular}}
  \footnotesize \textbf{*} Statistics for NLM180 and DDI2013 were computed on mapped examples (based on our own annotation mapping scheme) and not based on the original data.
\end{table*}

We train and evaluate our model on the Text Analysis Conference (TAC) 2018 dataset for drug-drug interaction extraction from drug labels~\citep{fushman2018overview}. 
The training data contains 22 drug labels, referred to as TR22, with gold standard annotations. 
As training data is scarce, we additionally propose a transfer learning step whereby the model is first trained on external data for extracting DDIs including the NLM-DDI CD corpus\footnote{\small https://lhce-brat.nlm.nih.gov/NLMDDICorpus.htm} and SemEval-2013 Task 9 dataset~\citep{herrero2013ddi}; we refer to these as NLM180 and DDI2013 respectively. Two official test sets of 57 and 66 drug labels, referred to as Test Set 1 and 2 respectively, with gold standard annotations are used strictly for evaluation. Table~\ref{tb_datastats} contains more information about these datasets and their characteristics. In this study, we show that the GCA improves over the standard GC and that our GCA based model with transfer learning by pretraining on external data improves over our the best model~\citep{tran2018multitask} from a prior study\footnote{\citet{tran2018multitask} was published as part of the non-refereed Text Analysis Conference (TAC); this study is an extension of our original report.}, that is based solely on BiLSTMs, by 4 absolute F1 points in overall performance. Furthermore, we show that our GCA based model complements our prior BiLSTM model; that is, by combining the two via ensembling, we improve over the prior best by 6 absolute F1 points in overall performance. Among comparable methods, our GCA based method exhibits state-of-the-art performance on all metrics after controlling for available training data. Our code\footnote{http://tttran.net/code/gcn-ddi2019.zip} is available for review and will be made publicly available on GitHub.

\section{Related Works}\label{sec-related}

Prior studies on DDI extraction have focused primarily on binary relation extraction where drug entities are known during test time and the learning objective is reduced to a simpler \emph{relation classification} (RC) task. In RC, pairs of known drug entities occurring in the same sentence are assigned a label, from a fixed set of labels, indicating relation type (including the \emph{none} or \emph{null} relation). Typically, no preliminary drug entity recognition or additional consequence prediction step is required. 
In this section, we cover prior relation extraction methods for DDI as well as participants of the initial TAC DDI challenge.


\subsection{Relation Extraction for DDI}

State-of-the-art methods for DDI extraction typically involve some variant of convolutional neural networks (CNNs) or recurrent neural networks (RNNs), or a hybrid of the two. Many studies utilize the dependency parse structure of an input sentence to capture long-distance dependencies, which has previously been shown to improve performance in general relation extraction tasks~\citep{zhang2018graph} and those in the biomedical domain~\citep{luo2016bridging,liu2016dependency}. \citet{liu2016drug} first proposed the use of standard CNNs for DDI extraction. Their approach involved convolving over an input sentence with drug entities bound to generic tokens in conjunction with so called \emph{position vectors}. Position vectors are used to indicate the offset between a word and each drug of the pair and provide additional spatial features. Improvements were attained, in a follow-up study, by instead convolving over the shortest dependency path between the candidate drug pair~\citep{liu2016dependency}. \citet{zhao2016drug} introduced an enhanced version of the CNN based method by deploying word embeddings that were pretrained on syntactic parses, part-of-speech embeddings, and traditional handcrafted features. \citet{suarez2017exploring} instead focused on fine-tuning various hyperparameter settings including word and position vector dimensions and convolution filter sizes for improved performance. \citet{kavuluru2017extracting} introduced the first neural architecture for DDI extraction based on hierarchical RNNs, wherein hidden intermediate representations are composed in a sequential fashion with cyclic connections, with character and word-level input. \citet{sahu2018drug} experimented with various ways of composing the output of a bidirectional LSTM network including max-pooling and attention pooling. \citet{lim2018drug} proposed a \emph{recursive} neural network architecture using recurrent units called TreeLSTMs to produce meaningful intermediate representations that are composed based on the structure of the dependency parse tree of a sentence. \citet{asada2018enhancing} demonstrated that combining representations of a CNN over the input text and graph convolutions over the molecular structure of the target drug pair (as informed by an external drug database) can result in improved DDI extraction performance. More recently, \citet{sun2019drug} proposed a hybrid RNN/CNN method by convolving over the contextual representations produced by a preceding recurrent neural network. 

\subsection{TAC 2018 DDI Track} 

TAC is a series of workshops organized by NIST aimed at encouraging research in natural language processing (NLP) by providing large test collections along with a standard evaluation procedure. The ``DDI Extraction from Drug Labels'' track~\citep{fushman2018overview}
is established with the goal of transforming the contents of drug labels into a machine-processable format with linkage to standard terminologies. \citet{tang2018two} placed first in the challenge using an encoder/decoder architecture to jointly identify precipitants and their interaction types and a rule-based system to determine interaction outcome. In addition to the provided training data, they downloaded and manually annotated a collection of 1148 sentences to be used as external training data. \citet{tran2018multitask} placed second in the challenge using a BiLSTM for joint entity recognition and interaction type prediction, followed by a CNN with two separate dense output layers (one of PK and one for PD) for outcome prediction. \citet{dandala2018ibm} placed third in the challenge using a BiLSTM (with CRFs) with part-of-speech and dependency features as input for entity recognition. Next, an Attention-LSTM model was used to detect relations between recognized entities. The embeddings were pretrained on a corpus of FDA-released drug labels and used to initialized the model. NLM180 was used for training with TR22 serving as the development set. Other participants proposed systems involving similar approaches including BiLSTMs and CNNs as well as traditional linear and rule-based methods.

\section{Materials and Methods}

We begin by formally describing the end-to-end task in Section~\ref{sec-task-description}. Next, we describe our approach to framing and modeling the problem (Section~\ref{sec-modeling-approach}), the proposed network architecture (Section~\ref{sec-nn}), the data used for transfer learning (Section~\ref{sec-datasets}), and our model-ensembling approach (Section~\ref{sec-ensembling}). Finally, in Section~\ref{sec-eval}, we describe the method for model evaluation. 

\subsection{Task Description}\label{sec-task-description}

Herein, we describe the end-to-end task of automatically detecting drugs and their interactions, including the outcome of identified interactions, as conveyed in drug labels.
We first define drug label as a collection of sections (e.g., \texttt{DOSAGE \& ADMINISTRATION}, \texttt{CONTRAINDICATIONS}, and \texttt{WARNINGS}) where each section contains one or more sentences. 
The overall task, in essence, involve fundamental language processing techniques including named entity recognition (NER) and relation extraction (RE). The first subtask of NER is focused on identifying \textbf{mentions} in the text corresponding to precipitants, interaction triggers, and interaction effects. Precipitating drugs (or simply precipitants) are defined as substances, drugs, or a drug class involved in an interaction. The second subtask of RE is focused on identifying sentence-level interactions; specifically, the goal is to identify the interacting precipitant, the type of the interaction, and outcome of the interaction. The interaction outcome depends on the interaction type as follows. Pharmacodynamic (PD) interactions are associated with a specified \emph{effect} corresponding to a span within the text that describes the outcome of the interaction. Figure~\ref{fig_tac_example} features a simple example of a PD interaction that is extracted from the drug label for \textbf{Adenocard}, where the precipitant is \emph{digitalis} and the effect is ``ventricular fibrillation.'' Naturally, it is possible for a precipitant to be involved in multiple PD interactions. Pharmacokinetic (PK) interactions, on the other hand, are associated with a label from a fixed vocabulary of National Cancer Institute (NCI) Thesaurus codes indicating various levels of increase/decrease in functional measurements. For example, consider the sentence: ``There is evidence that treatment with \texttt{phenytoin} leads to decrease intestinal absorption of \texttt{furosemide}, and consequently to lower peak serum \texttt{furosemide} concentrations.'' Here, \emph{phenytoin} is involved in a PK interaction with the label drug, \emph{furosemide}, and the type of PK interaction is indicated by the NCI Thesaurus code C54615 which describes a decrease in the maximum serum concentration (C$_{\text{max}}$) of the label drug. Lastly, \emph{unspecified} (UN) interactions  are interactions with an outcome that is not explicitly stated in the text and is typically indicated through cautionary remarks. 

\subsection{Joint Modeling Approach}\label{sec-modeling-approach}

\begin{table}[t]
	\caption{Example of the sequence labeling scheme for the sentence in Figure~\ref{fig_tac_example}, where \texttt{LABELDRUG} is substitute for Adenocard. \label{tb_example}\vspace{-1em}}
  \renewcommand{\arraystretch}{1.1}
  \begin{tabular}{@{\extracolsep{1em}}cccccccccc@{}}
  \toprule
  \texttt{O} & \texttt{O} & \texttt{O} & \texttt{O} & \texttt{O} & \texttt{O} & \texttt{O} & \texttt{\textbf{U-DYN}} \\
  The & use & of & LABELDRUG & in & patients & receiving & digitalis\\
  \midrule
  \texttt{O} & \texttt{O} & \texttt{O} & \texttt{\textbf{B-TRI}} & \texttt{\textbf{L-TRI}} & \texttt{\textbf{B-EFF}} & \texttt{\textbf{L-EFF}} & \texttt{O}\\
  may & be & rarely & associated & with & ventricular & fibrillation & .\\
  \bottomrule
  \end{tabular}
\end{table}

Since only precipitants are annotated in the ground truth, we model the task of precipitant recognition and interaction type prediction jointly. We accomplish this by reducing the problem to a sequence tagging problem via a novel NER tagging scheme. That is, for each precipitant drug, we additionally encode the associated interaction type. Hence, there are three possible precipitant tags: \textbf{DYN}, \textbf{KIN}, and \textbf{UN} for precipitants with pharmaco\textbf{dyn}amic, pharmaco\textbf{kin}etic, and \textbf{un}specified interactions respectively. Two more tags, \textbf{TRI} and \textbf{EFF}, are added to further identify mentions of \emph{triggers} and \emph{effects} concurrently. To properly identify boundaries, we employ the BILOU encoding scheme~\citep{ratinov2009design}. In the BILOU scheme, \emph{B}, \emph{I}, and \emph{L} tags are used to indicate the beginning, inside, and last token of a multi-token entity respectively. The \emph{U} tag is used for unit-length entities while the \emph{O} tag indicates that the token is outside of an entity span. As a preprocessing step, we identify the label drug in the sentence, if it is mentioned, and bind it to a generic entity token (e.g., ``LABELDRUG''). We also account for indirect mentions of the label drug, such as the generic version of a brand-name drug, or cases where the label drug is referred to by its drug class. To that end, we built a lexicon of drug names mapped to alias using NLM's Medical Subject Heading (MeSH) tree as a reference. Table~\ref{tb_example} shows how the tagging scheme is applied to a simple example.

Once we have identified the precipitant (as well as triggers/effects) and the interaction type for each precipitant, we subsequently predict the outcome or consequence of the interaction (if any). To that end, we consider all entity spans annotated with \textbf{KIN} tags and assign them a label from a static vocabulary of 20 NCI concept codes corresponding to PK consequence (i.e., multi-class classification). Likewise, we consider all entity spans annotated with \textbf{DYN} tags and link them to mention spans annotated with \textbf{EFF} tags; we accomplish this via binary classification of all pairwise combinations. For entity spans annotated with \textbf{UN} tags, no additional outcome prediction is needed.

\subsection{Notations and Neural Building Blocks}\label{sec-notation}

In this section, we describe notations used in the remainder of this study. In addition, we provide a generic definition of the canonical CNN and BiLSTM networks that are later used as building blocks in model construction. For ease of notation, we assume fixed sentence length $n$ and word length $\hat{n}$; in practice, we set $n$ and $\hat{n}$ to be the maximum sentence/word length and zero-pad shorter sentences/words. Moreover, we use square brackets with matrices to indicate a row indexing operation; for example, $\matrow{X}{i}$ denotes the vector corresponding to the \nth{i} row of matrix $X$. 

Henceforth, the abstract function $f_\text{CNN}^{w,d_\text{out}}(\cdot) : \mathbb{R}^{n \times d_\text{in}} \mapsto \mathbb{R}^{d_\text{out}}$ is used to represent the CNN that convolves on a window of size $w$ in a sentence with $n$ words, mapping an $n \times d_\text{in}$ matrix to a vector representation of length $d_\text{out}$, where $d_\text{in}$ is the word embedding size. This is an abstraction of the canonical CNN for NLP first proposed by~\citet{kimcnn} and is defined as follows. First, we denote the convolution operation $\star$ as the sum of the element-wise products of two matrices. That is, for two matrices A and B of same dimensions, $A \star B = \sum_{j} \sum_{k} A_{j,k} \cdot B_{j,k}$. Suppose the  input is a sequence of vector representations $\vec{x}^1, \ldots, \vec{x}^n \in \RR^{d_\text{in}}$; the output representation $\vec{g} \in \RR^{d_\text{out}}$ is defined such that
\begin{equation*}
\begin{split}
\vec{g}_k = \max ( &\, f_\text{convolve}(k,\vec{x}^{1}, \ldots, \vec{x}^{w}) \,, \, \ldots, \\
&\, f_\text{convolve}(k,\vec{x}^{n-w+1}, \ldots, \vec{x}^{n}) \, ) \\
& \text{ \hspace{1em} for } k = 1,\ldots,d_\text{out},\\
\end{split}
\end{equation*}
given a convolution function $f_\text{convolve}$ that \emph{convolves} over a contiguous window of size $w \leq n$, defined as  
\begin{equation*}
f_\text{convolve}(k,\vec{v}^1, \ldots, \vec{v}^w) = 
\text{ReLU} \left( W^k \star 
\left(\begin{array}{c}
\vec{v}^1\\
\vdots\\
\vec{v}^w\\
\end{array}\right) + b^k\right)
\end{equation*}
where $\vec{v}^1, \ldots, \vec{v}^w \in \RR^{d_\text{in}}$ are input vectors, $W^k \in \RR^{w \times d_\text{in}}$ and $b^k \in \RR$ for $k = 1,\ldots,d_\text{out}$, are network parameters (corresponding to a set of $d_\text{out}$ convolutional filters), $\text{ReLU(x)} = \max(0,x)$ is the linear rectifier activation function. Here, $d_\text{out}$ is a hyperparameter that determines the number of convolutional filters and thus the size of the final feature vector. In the study, we denote the convolution as an abstract function $f_\text{CNN}^{w,d_\text{out}}(\cdot) : \mathbb{R}^{n \times d_\text{in}} \mapsto \mathbb{R}^{d_\text{out}}$ that convolves on a window of size $w$ and maps an $n \times d_\text{in}$ matrix to a vector representation of length $d_\text{out}$.

Likewise, we represent the BiLSTM network as an abstract function  $f^{d_\text{out}}_\text{BLSTM}(\cdot) : \mathbb{R}^{n \times d_\text{in}} \mapsto \mathbb{R}^{n \times d_\text{out}}$ that maps a sequence of $n$ input vectors (e.g., word embeddings) of $d_\text{in}$ size (as an $n \times d_\text{in}$ matrix) to a corresponding sequence of $n$ output context vectors of $d_\text{out}$ size (as an $n \times d_\text{out}$ matrix). 
Let $\LSTMR$ and $ \LSTML$ represent an LSTM composition in the forward and backward direction. Suppose the  input is a sequence of vector representations $\vec{x}^1, \ldots, \vec{x}^n \in \RR^{d_\text{in}}$; the output of a standard bidirectional LSTM network (BiLSTM) is a matrix $H \in \RR^{n \times d_\text{out}} = \left( \vec{h}^1, \ldots, \vec{h}^n \right)^{\top}$ such that 
\begin{align*}
  \lvec{h}^i &= \LSTMR (\vec{x}^i), \,\,\\
  \rvec{h}^i &= \LSTML (\vec{x}^i), \\
  \vec{h}^i &= \lvec{h}^i  \concat \rvec{h}^i,
  \text{ \hspace{1em} for } i = 1,\ldots,n,
\end{align*} 
where $\concat$ is the vector concatenation operator and $\vec{h}^i \in \RR^{d_\text{out}}$ represents the context centered at the \nth{i} word. Here, $d_\text{out}$ is a hyperparameter that determines the size of the the context embeddings. In the study, we denote the BiLSTM network as an abstract function  $f^{d_\text{out}}_\text{BLSTM}(\cdot) : \mathbb{R}^{n \times d_\text{in}} \mapsto \mathbb{R}^{n \times d_\text{out}}$ that maps a sequence of $n$ input vectors (e.g., word embeddings) of $d_\text{in}$ size (as an $n \times d_\text{in}$ matrix) to a corresponding sequence of $n$ output context vectors of $d_\text{out}$ size (as an $n \times d_\text{out}$ matrix).

\subsection{Neural Network Architecture and Training Details}\label{sec-nn}

\begin{figure*}[t]
  \center{\includegraphics[width=\textwidth]
  {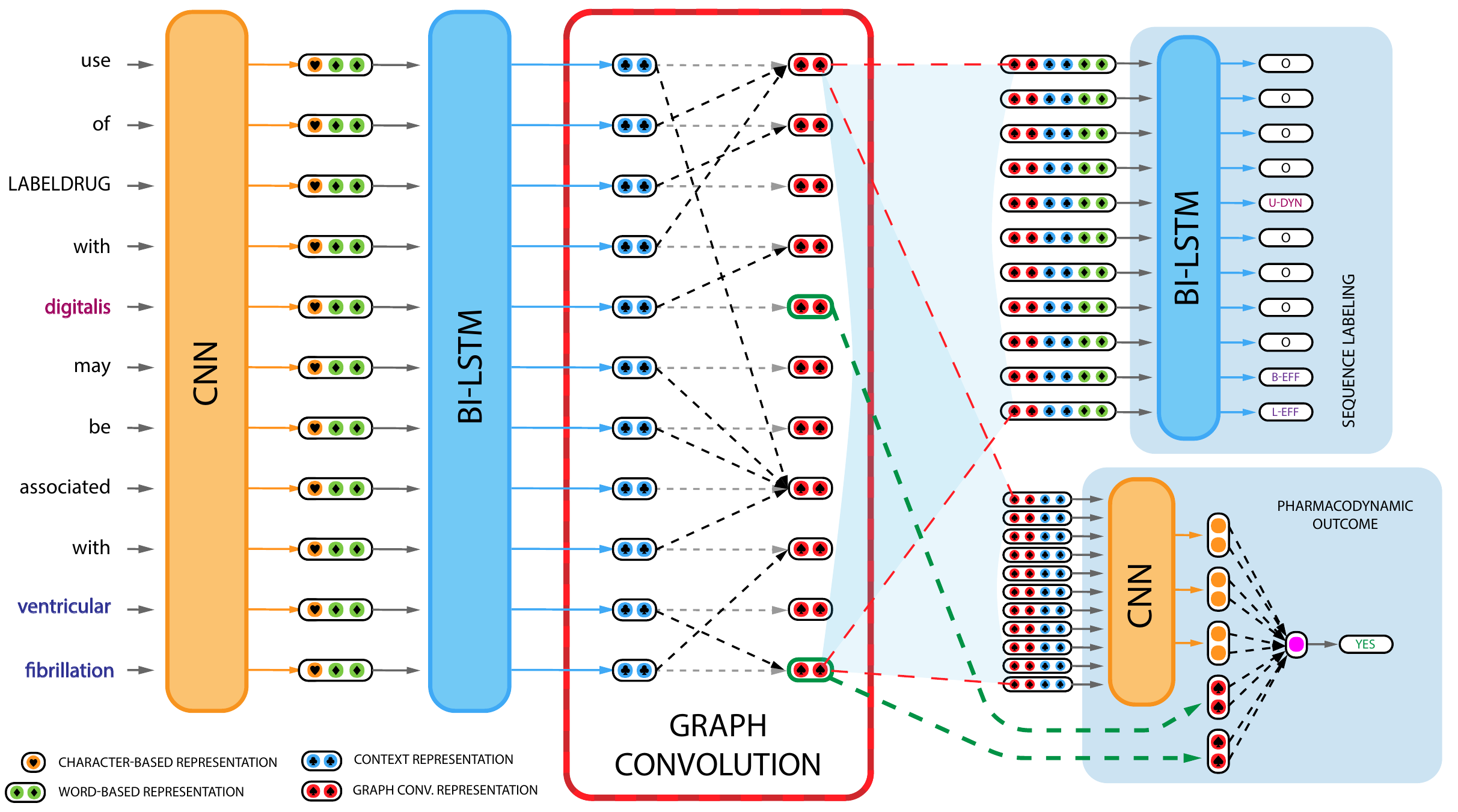}}
  \caption{Overview of the neural network architecture for a simplified example from the drug label Adenocard. Here, the ground truth indicates that \emph{digitalis} is a pharmacodynamic precipitant associated with the effect ``ventricular fibrillation.'' The PK predictive component is omitted given there are no precipitants involved in a PK interaction. }
  \label{fig_model_v2}
\end{figure*}

We begin by describing how the three types of intermediate representations are composed. The construction of word, context, and graph-based representations are described in Sections~\ref{sec-word-rep}, \ref{sec-context-rep}, and \ref{sec-graph-rep} respectively. Next, we describe the predictive components of the network that share and utilize the intermediate representations. In Section~\ref{sec-seq-lab}, we describe the sequence-labeling component of the network used to extract drugs and their interactions. In Section~\ref{sec-consequence-pred}, we describe the component for predicting interaction outcome. An overview of the architecture is shown in Figure~\ref{fig_model_v2}. Lastly, we describe the model configuration and training process in Section~\ref{sec-training}.

\subsubsection{Word-level Representation}\label{sec-word-rep}
Suppose the input is a sentence of length $n$ represented by a sequence of word indices $w_1, \ldots, w_n$ into the vocabulary $\set{V}^{\text{Word}}$. Each word is mapped to a word embedding vector via embedding matrices $E^{\text{Word}} \in \RR^{|\set{V}^{\text{Word}}| \times \delta}$ such that $\delta$ is a hyperparameter that determines the size of word embeddings. 
In addition to word embeddings, we employ character-CNN based representations as commonly observed in recent neural NER models~\citep{chiu2016named}. Character-based models capture morphological features and help generalize to out-of-vocabulary words. 
For the proposed model, such representations are composed by convolving over character embeddings of size $\pi$ using a window of size 3, producing $\eta$ feature maps; the feature maps are then max-pooled to produce $\eta$-length feature representations. Correspondingly, we denote  $E^{\text{Char}} \in \RR^{|\set{V}^{\text{Char}}| \times \pi}$ as the embedding matrix given the character vocabulary $\set{V}^{\text{Char}}$; the character-level embedding matrix $C^i \in \RR^{\hat{n} \times \pi}$ for the word at position $i$ is 
\begin{equation*}
C^i = \left(
\begin{array}{c}
\matrow{E^{\text{Char}}}{c_{i,1}} \\
\vdots\\
\matrow{E^{\text{Char}}}{c_{i,\hat{n}}} \\
\end{array}
\right) \\
\end{equation*}
where 
$c_{i,j}$ for $1 \leq i \leq n, 1 \leq j \leq \hat{n},$ represents the \nth{j} character index of the \nth{i} word. The word-level representation $R^{\text{word}} \in \RR^{n \times (\delta + \eta)}$ is a concatenation of character-based word embeddings and pretrained word embeddings along the feature dimension; formally, 
\begin{equation*}
R^{\text{Word}} = \left(
\begin{array}{c}
\matrow{E^{\text{Word}}}{w_1} \concat f_\text{CNN}^{3,\eta}(C^1) \\
\vdots\\
\matrow{E^{\text{Word}}}{w_n} \concat f_\text{CNN}^{3,\eta}(C^n) \\
\end{array}
\right). \\
\end{equation*}

\subsubsection{Context-based Representation}\label{sec-context-rep}
We compose context-based representation by simply processing the word-level representation with a BiLSTM layer as is common practice; concretely, $R^{\text{Context}} = f^{\rho}_\text{BLSTM}(R^{\text{Word}})$ where $\rho$ is a hyperparameter that determines the size of the context embeddings.

\subsubsection{Graph-based Representation}\label{sec-graph-rep}
In addition to the sequential nature of LSTMs, we propose an alternative and complementary graph-based approach for representing context using graph convolution (GC) networks. Typically composed on dependency parse trees, graph-based representations are useful for relation extraction as they capture long-distance relationships among words of a sentence as informed by the sentence's syntactic dependency structure. While graph convolutions are typically applied repeatedly, our initial cross-validation results indicate that single-layered GCs are sufficient and deep GCs typically resulted in performance degradation; moreover, \citet{zhang2018graph} report good performance with similarly shallow GC layers. Hence the following formulation describes a \emph{single-layered} GC network, with an additional attention-based sigmoid gating mechanism, which we holistically refer to as a Graph Convolution with Attention-Gating (GCA) network. Initially motivated in Section~\ref{sec-intro}, the GCA improves on conventional GCs with a sigmoid-gating mechanism derived via an alignment score function associated with \emph{additive} attention~\citep{bahdanau2014neural}. The sigmoid ``gate'' determines whether or not (and to what extent) information is propagated based on a learned alignment function that conceives a ``relevance'' score between a \emph{source} and \emph{target} node (more later). 

As a pre-processing step, we use a dependency parsing tool to generate the projective dependency tree for the input sentence. We represent the dependency tree as an $n \times n$ adjacency matrix $A$ where $A_{i,j} = A_{j,i} = 1$ if there is a dependency relation between words at positions $i$ and $j$. This matrix controls the flow of information between pairs of words corresponding to connected nodes in the dependency tree (ignoring dependency \emph{type}); however, it is also important for the \emph{existing} information of each node to carry over on each application of the GC. Hence, as with prior work~\citep{zhang2018graph}, we use the modified version $\tilde{A} = A + I$ where $I$ is the identity matrix to allow for self-loops in the GC network. The graph-based representation $R^{\text{Graph}} \in \RR^{n \times \beta}$ is composed such that 
\begin{equation*}
\matrow{R^{\text{Graph}}}{i} = \tanh \left( \sum^{n}_{j=1} \tilde{A}_{i,j} W^{\text{Graph}} \matrow{R^{\text{Context}}}{j} + \vec{b}^{\text{Graph}} \right)
\end{equation*}
where $W^{\text{Graph}} \in \RR^{\beta \times \rho}, \vec{b}^{\text{Graph}} \in \RR^{\beta}$ are network parameters, $\tanh(\cdot)$ is the hyperbolic tangent activation function, and $\beta$ is a hyperparameter that determines the hidden GC layer size. Thus, information propagated from source nodes $j = 1,\ldots,n$ to target node $i$, based on the summation of intermediate representations, are unweighted and share equal importance.

As stated previously, we propose to extend the standard GC by adding an attention-based sigmoid gating mechanism to control the flow of information via the gating matrix $G \in \RR^{n \times n}$. We define $G$ such that 
\begin{equation*}
G_{i,j} = \sigma ( \vec{v} \cdot \vec{a}^{i,j} ) \text{ \hspace{1em} for } i = 1,\ldots,n, j = 1,\ldots,n,
\end{equation*}
where $\vec{v} \in \RR^{\alpha}$ is a network parameter and $\vec{a}^{i,j} \in \RR^{\alpha}$ is the hidden attention layer composed as a function of the context representation at source node $i$ and target node $j$; concretely, 
\begin{equation*}
\vec{a}^{i,j} = \tanh \left( W^{\text{Source}} \matrow{R^{\text{Context}}}{i} + W^{\text{Target}} \matrow{R^{\text{Context}}}{j} + \vec{b}^{\text{Attn}} \right) \,, 
\end{equation*}
where $W^{\text{Source}}, W^{\text{Target}} \in \RR^{\alpha \times \rho}$ and $\vec{b}^{\text{Attn}} \in \RR^{\alpha}$ are network parameters and $\alpha$ is a hyperparameter that determines hidden attention layer size. Intuitively, the network learns the relevance of node $i$ to node $j$ via the attention $\vec{a}^{i,j}$ and outputs a between 0 and 1 at gate $G_{i,j}$. Gate $G_{i,j}$ controls the flow of information from node $i$ to $j$, where 0 indicates no information is passed and 1 indicates that all information is passed. To integrate the gating mechanism, we simply redefine $\tilde{A} = (A + I) \times G$. In the next two sections, we show how the intermediate representations are used for end-task prediction.

\subsubsection{Sequence Labeling}\label{sec-seq-lab}

The sequence labeling (SL) task for detecting precipitant drugs and their interaction type is handled by a bidirectional LSTM trained on a combination of the two types of losses: conditional random fields (CRF) and softmax cross entropy (SCE).  Using CRFs results in choosing a globally optimal assignment of tags to the sequence, whereas a standard softmax at the output of each step may result in less globally consistent assignments (e.g., an L tag following an O tag) but better local or partial assignments. We begin by introducing a bidirectional LSTM layer that processes the various intermediate representations. The new representation, $R^{\text{SL}} \in \RR^{n \times \gamma}$, is defined such that 
\begin{equation*}
R^{\text{SL}} = f^{\gamma}_\text{BLSTM} \left(
\begin{array}{c}
\matrow{R^{\text{Word}}}{1} \concat \matrow{R^{\text{Context}}}{1} \concat \matrow{R^{\text{Graph}}}{1}\\
\vdots\\
\matrow{R^{\text{Word}}}{n} \concat \matrow{R^{\text{Context}}}{n} \concat \matrow{R^{\text{Graph}}}{n}\\
\end{array}
\right) \\
\end{equation*}
where $\gamma$ is a hyperparameter that determines the hidden layer size. While $R^{\text{Graph}}$ is based on $R^{\text{Context}}$ and $R^{\text{Context}}$ is based on $R^{\text{Word}}$, we observed that combining these intermediate representations (manifesting at varying depth in the architecture) resulted in improved sequence-labeling performance according to preliminary experiments and prior results from~\citet{tran2018multitask}. As with residual networks~\citep{he2016deep}, they additionally provide a kind of shortcut or ``skip-connection'' over intermediate layers.

Given a set of $n_{\text{tag}}$ possible tags, we compose an $n \times n_{\text{tag}}$ score matrix $Y$ (where $Y_{i,t}$ represents the score of the \nth{t} tag at position $i$) such that  
$\matrow{Y}{i} = W^{\text{Out}}\matrow{R^{\text{SL}}}{i} + \vec{b}^{\text{Out}}$
where 
$W^{Out} \in \RR^{n_{\text{tag}} \times \gamma}, \vec{b}^{\text{Out}} \in \RR^{n_{\text{tag}}}$ are network parameters. Given example $x$ and the truth tag assignment as a matrix $\bar{Y}$ where rows are one-hot vectors over all possible tags, the SCE loss is 
\begin{equation*}
\ell_{\text{SCE}} (x, \tilde{A}, \bar{Y}; \theta) = -\sum^{n}_{i=1} \sum^{n_{\text{tags}}}_{t=1} \bar{Y}_{i,t} \log \left( \frac{ \exp (Y_{i,t}) }{ \sum^{n_{\text{tags}}}_{k=1} \exp( Y_{i,k}) } \right)
\end{equation*}
where $\bar{Y}_{i,t} \in \{0,1\}$ indicates whether the tag $t$ is assigned at position $i$ and $\theta$ is the set of all network parameters.
Next, we define the CRF loss as commonly used with LSTM based models for entity recognition. We learn a transition score matrix $M \in \RR^{n_{\text{tag}} \times n_{\text{tag}}}$, inferred from the training data, such that $M_{i,j}$ is the transition score from tag $i$ to tag $j$. 
Given an example $x$ as a sequence of word indices $w_1, \ldots, w_n$ and candidate tag sequence $\bar{y}$ as a sequence of tag indices $s_1, \ldots, s_n$, the tag assignment score (t-score) is defined as 
\begin{align*}
\text{t-score} \, ( x, \tilde{A}, \bar{y}; \hat{\theta}) &= \text{t-score} \, ( w_1, \ldots, w_n, \tilde{A}, s_1, \ldots, s_n; \hat{\theta}) = \sum^{n}_{i=1} \left( Y_{i,s_{i}} + M_{s_{i-1},s_{i}} \right)
\end{align*}
where $\hat{\theta} = \theta \cup \{ M \}$. Intuitively, this score summarizes the likelihood of observing a transition from tag $s_{i-1}$ to tag $s_{i}$ in addition to the likelihood of emitting tag $s_i$ given the semantic context for $i = 1, \ldots, n$. Thus $Y$ is treated as a matrix of emission scores for the CRF. For an example with input $x$ and truth tag assignment $\bar{y}$, the loss is computed as the negative log-likelihood of the tag assignment as informed by the normalized tag assignment score, or 
\begin{equation*}
\ell_{\text{CRF}} (x, \tilde{A}, \bar{y}; \hat{\theta}) = - \log \frac{\exp (\text{t-score} (x, \tilde{A}, \bar{y}; \hat{\theta})) }{\sum_{y \in S} \exp (\text{t-score} (x, \tilde{A}, y; \hat{\theta}) )}
\end{equation*}
where $S$ is the set of all possible tag assignments. The final per-example loss for sequence labeling is simply a summation of the two losses: $\ell_{\text{SL}} = \ell_{\text{SCE}} + \ell_{\text{CRF}}$. During testing, we use the Viterbi algorithm~\citep{viterbi1967error}, a dynamic programming approach, to decode and identify the globally optimal tag assignment.

\subsubsection{Consequence prediction}\label{sec-consequence-pred}
 
Once precipitants (and corresponding interaction types) have been identified, we perform so called consequence prediction (CP) for all precipitant drugs identified as participating in PD or PK interactions. The classification task of CP takes as input the target sentence and two candidate entities that are referred to as the \emph{subject} and \emph{object} entities. Here the \emph{subject} is always a precipitating drug; on the other hand, the \emph{object} designation depends on the type of interaction (more later). First, we define the representation matrix for CP as $R^{\text{CP}} \in \RR^{n \times (\rho + \beta)}$ where 
\begin{equation*}
R^{\text{CP}} = \left(
\begin{array}{c}
\matrow{R^{\text{Context}}}{1} \concat \matrow{R^{\text{Graph}}}{1}\\
\vdots\\
\matrow{R^{\text{Context}}}{n} \concat \matrow{R^{\text{Graph}}}{n}\\
\end{array}
\right). \\
\end{equation*}
We process the matrix via convolutions of windows sizes 3, 4, and 5 and concatenate the results to produce the final feature vector $\vec{g}^\text{CP}$. In addition to CNN features, we map entities to their graph based context features and append it to $\vec{g}^\text{CP}$, which has been previously shown to work well in a similar architecture~\citep{li2017neural}. Concretely, the final feature vector is
\begin{equation*}
\vec{g}^\text{CP} = f_\text{CNN}^{3,\mu} (R^{\text{CP}}) 
\concat f_\text{CNN}^{4,\mu} (R^{\text{CP}}) 
\concat f_\text{CNN}^{5,\mu} (R^{\text{CP}}) 
\concat R^{\text{CP}}[t_{\text{Sub}}] 
\concat R^{\text{CP}}[t_{\text{Obj}}] 
\end{equation*}
with $\vec{g}^\text{CP} \in \RR^{3\mu + 2(\rho + \beta)}$ where $\mu$, as a hyperparameter, is the number of CNN filters per convolution
and $t_{\text{Sub}}$ and $t_{\text{Obj}}$ are the position index of the last word (typically the ``head'' word) of the subject and object respectively.

The actual entities determined to be the subject/object pair are based on the interaction type; for PD interactions, the \emph{subject} is the precipitant drug and the \emph{object} is some candidate \emph{effect} mention. For PK interactions, however, the subject is the precipitant drug but the object is chosen to be the closest (based on character-offset) mention of the drug label with respect to the target precipitant drug. We found this appropriate based on manual review of the data, as the NCI code being assigned depends highly on whether the increase/decrease in functional measurements is with respect to the label drug or the precipitant drug. In case the label drug is not mentioned, a generic ``null'' vector is used to represent the object.

When performing sequence labeling, we pass in the entire dependency tree encoded as the matrix $\tilde{A}$. However, when performing consequence prediction and both entities are non-null, we pass in a \emph{pruned} version of the entire tree that is tailored to the entity pair. We apply the same pruning strategy proposed by \citet{zhang2018graph}, wherein for a pair of subject and object entities (corresponding to $t_{\text{Sub}}$ and $t_{\text{Obj}}$), we keep only nodes either along or \emph{within one hop} of the shortest dependency path. This prevents distant and irrelevant portions of the dependency tree from influencing the model while retaining important modifying and negating terms. Thus the notation $\ASO$ is used to denote the pruned version of $\tilde{A}$ as a function of the entity pair indicated by $t_{\text{Sub}}$ and $t_{\text{Obj}}$.

To determine whether there is a PD interaction between a pair of entities, we employ a standard binary classification output layer. Concretely, for example sentence $\hat{x}$ and output $y \in \{0,1\}$, the probability of a PD interaction between the entity pair is 
$q = \text{sigmoid} (\vec{w}^{\text{PD}} \cdot \vec{g}^\text{CP} + b^{\text{PD}})$
where $\vec{w}^{\text{PD}} \in \RR^{3\mu + 2(\rho + \beta)}$ and $b^{\text{PD}} \in \RR$ are network parameters. The associated binary cross entropy loss is 
\begin{equation*}
\ell_{\text{PD}} (x, \ASO, \hat{y}; \theta) = \hat{y} \log q + (1-\hat{y}) \log (1-q)
\end{equation*}
where $\hat{y} \in \{0,1\}$ indicates the ground truth.
For PK interactions, we instead use a softmax function to produce a probability distribution, represented as vector $\vec{q} \in \RR^{20}$, over the 20 labels corresponding to NCI Thesaurus codes. Concretely, the predicted probability of label $j$ is 
$\vec{q}_j = \exp (\vec{y}^{\text{PK}}_j ) / \exp( \sum^{20}_{k=1}\vec{y}^{\text{PK}}_k )$ where 
$\vec{y}^{\text{PK}} = W^{\text{PK}} \vec{g}^\text{CP} + \vec{b}^{\text{PK}}$ and $W^{\text{PK}} \in \RR^{20 \times [3\mu + 2(\rho + \beta)]}$ and $\vec{b}^{\text{PK}} \in \RR^{20}$ are network parameters. Given a one-hot vector $\bar{\vec{y}} \in \RR^{20}$ indicating the ground truth, the associated softmax cross entropy loss is 
\begin{equation*}
\ell_{\text{PK}} (x, \ASO, \bar{\vec{y}}; \theta) = \sum^{20}_{j=1} \bar{\vec{y}}_j \log \vec{q}_j \quad .
\end{equation*} 
The loss for a batch of examples is simply the sum of its constituent example-based losses. 

\subsubsection{Neural Network Configuration and Training Details}\label{sec-training}

\begin{table}[!t]
		\caption{Model configuration obtained through random search over 11-fold cross-validation of TR22 (training data).\label{tab_config}}
    \renewcommand{\arraystretch}{1.1}
    \begin{tabular}{cc}
    \toprule
    \begin{tabular}{@{\extracolsep{1em}}lr}
    \textbf{Setting} & \textbf{Value}\\
    \midrule
    Learning Rate & 0.001 \\
    Dropout Rate & 0.5\\
    Character Embedding Size ($\pi$) \hspace{1em} & 25\\ 
    Character Representation Size ($\eta$) & 50\\ 
    Word Embedding Size ($\delta$) & 200\\
    \end{tabular} & 
    \begin{tabular}{@{\extracolsep{2em}}lr}
    \textbf{Setting} & \textbf{Value}\\
    \midrule
    Context Embedding Size ($\rho$) & 100\\
    GC Hidden Size ($\beta$) & 100\\ 
    GC Attention Size ($\alpha$) & 25\\ 
    Sequence LSTM Hidden Size ($\gamma$) & 200 \\
    Outcome CNN Filter Count ($\mu$) & 50\\ 
    \end{tabular}\\
    \bottomrule
    \end{tabular}
\end{table}

For each training iteration, we randomly sample 10 sentences from the training data. These are re-composed into three sets of task-specific examples $\mathcal{S}$, $\mathcal{D}$, and $\mathcal{K}$ corresponding to the tasks of sequence labeling, PD prediction, and PK prediction respectively. Unlike our prior work, in which the sub-tasks were trained in an interleaved fashion, we train on all three objectives jointly. Here, we dynamically switch between one of four training objective losses based on whether there are available training examples (in the batch and for the current iteration) for each task. The final training loss is then 
\begin{equation*}
\ell = \begin{cases} 
        \,\sum\limits_{x \in \mathcal{S}}\ell_{\text{SL}}(x) + \sum\limits_{x \in \mathcal{D}}\ell_{\text{PD}}(x) + \sum\limits_{x \in \mathcal{K}}\ell_{\text{PK}}(x) \quad\quad & \text{if } |\mathcal{D}| > 0 \text{ and } |\mathcal{K}| > 0,\\
        \,\sum\limits_{x \in \mathcal{S}}\ell_{\text{SL}}(x) + \sum\limits_{x \in \mathcal{K}}\ell_{\text{PK}}(x) & \text{if } |\mathcal{D}| = 0 \text{ and } |\mathcal{K}| > 0,\\
        \,\sum\limits_{x \in \mathcal{S}}\ell_{\text{SL}}(x) + \sum\limits_{x \in \mathcal{D}}\ell_{\text{PD}}(x) & \text{if } |\mathcal{D}| > 0 \text{ and } |\mathcal{K}| = 0,\\
        \,\sum\limits_{x \in \mathcal{S}}\ell_{\text{SL}}(x) & \text{otherwise}.\\
        \end{cases}
\end{equation*}

We train the network for a maximum of 10,000 iterations, check-pointing and evaluating every 100 iterations on a validation set of sentences from four held-out drug labels. Only the checkpoint that performed best on the validation set is kept for test time evaluation. The choice of hyperparameters is shown in Table~\ref{tab_config}; discrete numbered parameters corresponding to embedding or hidden size were chosen from \{10, 25, 50, 100, 200, 400\} based on random search and optimized by assessing 11-fold cross-validation performance on TR22. The learning and dropout rates are set to typical default values. We used Word2Vec embeddings pretrained on the corpus of PubMed abstracts~\citep{pyysalo2013distributional}. All other variables are initialized using values drawn from a normal distribution with a mean of $0$ and standard deviation of $0.1$ and further tuned during training. Words were tokenized on both spaces and punctuation marks; punctuation tokens were kept as is common practice for NER type systems. For dependency parsing, we use SyntaxNet\footnote{https://github.com/tensorflow/models/tree/master/research/syntaxnet} which implements the transition-based neural model by \citet{andor2016globally}. We trained the aforementioned parser, using default settings, on the GENIA corpus~\citep{kim2003genia} and use it to obtain projective dependency parses for each example. 


\subsection{Transfer Learning with Network Pre-Training}\label{sec-datasets}

An obstacle in solving this flavor of DDI extraction as a machine learning problem is the high potential for overfitting given the sparse nature of the output space, which is further intensified by the scarce availability of high quality training data. As quality training data is expensive and requires domain expertise, we propose to use a transfer learning approach where the model is pre-trained on external data as follows. First, we pre-train on the DDI2013 dataset, which contains strictly binary relation DDI annotations and no interaction consequence annotation. Hence, DDI2013 is only used to train the sequence labeling objective $\ell_{\text{SL}}(x)$. Next, we pre-train on NLM180, a collection of 180 drug labels annotated in a comparable format to TR22 but  follows a different set of guidelines and lacks comprehensive interaction consequence annotation. Finally, we fine-tune for the target task by training on the official TR22 dataset. 

Translating NLM180 and DD2013 to the TAC 2018 format is an imperfect process given structural (breadth and depth of annotations) and semantic (guidelines in addition to annotator experience and vision) differences. For example, differences in how entity boundaries are annotated, such as whether or not modifier terms should be kept as part of a named entity, may have a large impact on model performance. Hence, we expect the translated versions of NLM180 and DDI2013 to be very noisy as training examples for the target task. We describe the translation process for  DDI2013 in Sections~\ref{sec-nlm180} and~\ref{sec-ddi2013}. We provide summary statistics about these datasets in Table~\ref{tb_datastats}.

\subsubsection{NLM180 Mapping Scheme}\label{sec-nlm180}
In NLM180, there is no distinction between triggers and effects; moreover, PK effects are limited to \emph{coarse}-grained (binary) labels corresponding to \emph{increase} or \emph{decrease} in function measurements. Hence, a direct mapping from NLM180 to the TR22 annotation scheme is impossible. As a compromise, NLM180 ``triggers'' were mapped to TR22 \emph{triggers} in the case of unspecified and PK interactions. For PD interactions, we instead mapped NLM180 ``triggers'' to TR22 \emph{effects}, which we believe to be appropriate based on our manual analysis of the data. Since we do not have both \emph{trigger} and \emph{effect} for every PD interaction, we opted to ignore trigger mentions altogether in the case of PD interactions to avoid introducing mixed signals. While trigger recognition has no bearing on relation extraction performance, this policy has the effect of reducing the recall upperbound on NER by about 25\% based on early cross-validation results. To overcome the lack of fine-grained annotations for PK outcome in NLM180, we deploy the well-known bootstrapping approach~\citep{jones1999bootstrapping} to incrementally annotate NLM180 PK outcomes using TR22 annotations as a starting point. To mitigate the problem of semantic drift, we re-annotated by hand iterative predictions that were not consistent with the original NLM180 coarse annotations (i.e., active learning~\citep{settles2012active}).

\subsubsection{DDI2013 Mapping Scheme}\label{sec-ddi2013}

The DDI2013 dataset contains annotations that are incomplete with respect to the target task; specifically, annotations are limited to typed binary relations between any two drug mentioned drugs in the sentence (and not necessary between a mentioned drug and the label drug) without outcome or consequence prediction. In DDI2013, there are four types of interactions: \emph{mechanism}, \emph{effect}, \emph{advice} and \emph{int}. The \emph{mechanism} type indicates that a PK mechanism is being discussed; \emph{effect} indicates that the consequence of a PD interaction is being discussed; \emph{advice} indicates suggestions regarding the handling of the drugs; and \emph{int} is an interaction without any specific additional information. We translate the annotation by first applying a filtering step on all interactions such that it conforms to the target task; namely, we filter such that only interactions involving the label drug is kept. The non-label drug entity is then annotated as a precipitant with an interaction tag based on the following mapping scheme. Entities involved in a \emph{mechanism} relation with the drug label are treated as \textbf{KIN} precipitants; likewise, entities in \emph{effect} and \emph{advice} relations are treated as \textbf{DYN} precipitants and \emph{int} relations are treated as \textbf{UNK} precipitants. As there is no consequence annotation, the mapped examples are  used to train the sequence labeling objective but \emph{not} the other objective.

\subsection{Voting-based Ensembling}\label{sec-ensembling}

Our prior effort~\citep{tran2018multitask} showed that model ensembling resulted in optimal performance for this task. Hence, model ensembling remains a key component of the proposed model. Our ensembling method is based on ensembling over \emph{ten} models each trained with randomly initialized weights and a random development split. Intuitively, models collectively ``vote'' on predicted annotations that are kept and annotations that are discarded. A unique annotation (entity or relation) has one vote for each time it appears in one of the \emph{ten} model prediction sets. In terms of implementation, unique annotations are incrementally added (to the final prediction set) in order of descending vote count; subsequent annotations that conflict (i.e., overlap based on character offsets) with existing annotations are discarded. Hence, we loosely refer to this approach as ``voting-based'' ensembling.\vspace{-0.1em}

\subsection{Model Evaluation}\label{sec-eval}

We used the official evaluation metrics for NER and relation extraction based on the standard precision, recall, and F1 micro-averaged over exactly matched  entity/relation annotations. We use the strictest matching criteria corresponding to the official ``primary'' metric (of the TAC DDI task), as opposed to the ``relaxed'' metric that ignores mention and interaction \emph{type}. Concretely, the matching criteria for entity recognition considers entity bounds as well as the type of the entity. The matching criteria for relation extraction comprehensively considers precipitant drugs and, for each, the corresponding interaction type and interaction outcome. As relation extraction evaluation takes into account the bounds of constituent entity predictions, relation extraction performance is heavily reliant on entity recognition performance. On the other hand, we note that while NER evaluation considers \emph{trigger} mentions, \emph{triggers} are ignored when evaluating relation extraction performance. Two test sets of 57 and 66 drug labels, referred to as Test Set 1 and 2 respectively, with gold standard annotations are used for evaluation.

Next, we discuss the differences between these test sets. As shown in Table~\ref{tb_datastats}, Test Set 1 closely resembles TR22 with respect to the sections that are annotated. However, Test Set 1 is more sparse in the sense that there are more sentences per drug label (144 vs. 27), with a smaller proportion of those sentences having gold annotations (23\% vs. 51\%). Test Set 2 is unique in that it contains annotations from only two sections, namely DRUG INTERACTIONS and CLINICAL PHARMACOLOGY, the latter of which is not represented in TR22 (nor Test Set 1). Lastly, TR22, Test Set 1, and Test Set 2 all vary with respect to the distribution of interaction types, with TR22, Test Set 1, and Test Set 2 containing a higher proportion of PD, UN, and PK interactions respectively. Overall model performance is assessed using a single metric defined as the average of \emph{entity recognition} and \emph{relation extraction} performance across both test sets.

\section{Results and Discussion}

\begin{table*}[!t]
	\caption{Main results based on 95\% confidence interval around mean precision, recall, and F1 based on evaluating N=100 ensembles for each model.\label{tb_results}}
  \renewcommand{\arraystretch}{1.8}
  \resizebox{\textwidth}{!}{
  \begin{tabular}{@{\extracolsep{0em}}lc ccc ccc ccc ccc ccc}
    \toprule
     \, & \, & \multicolumn{3}{c}{Test 1 / Entity} & \multicolumn{3}{c}{Test 1 / Relation} & \multicolumn{3}{c}{Test 2 / Entity} & \multicolumn{3}{c}{Test 2 / Relation} & \multicolumn{3}{c}{Overall}\\
     \cline{3-5}\cline{6-8}\cline{9-11}\cline{12-14}\cline{15-17}
     \textbf{Method} & \textbf{Training Data} & \textbf{P} (\%) & \textbf{R} (\%) & \textbf{F} (\%) & \textbf{P} (\%) & \textbf{R} (\%) & \textbf{F} (\%) & \textbf{P} (\%) & \textbf{R} (\%) & \textbf{F} (\%) & \textbf{P} (\%) & \textbf{R} (\%) & \textbf{F} (\%) & \textbf{P} (\%) & \textbf{R} (\%) & \textbf{F} (\%) \\
\midrule
BL & TR22 
& 23.82 & 42.04 & 30.39 & 14.74 & 18.38 & 16.35 & 26.15 & 39.69 & 31.51 & 12.48 & 15.43 & 13.79 & 19.30 $\pm$ 0.12 & 28.88 $\pm$ 0.12 & 23.01 $\pm$ 0.06\\
GCA & TR22 
& 32.87 & 32.35 & 32.59 & 22.70 & 13.95 & 17.27 & 38.82 & 31.31 & 34.65 & 19.26 & 11.63 & 14.49 & 28.41 $\pm$ 0.09 & 22.31 $\pm$ 0.16 & 24.75 $\pm$ 0.11\\
\midrule
BL$^{(1)}$ & TR22 + NLM180 
& 27.05 & 39.87 & 32.22 & 19.94 & 22.20 & 21.00 & 32.49 & 41.92 & 36.60 & 21.82 & 23.93 & 22.82 & 25.32 $\pm$ 0.09 & 31.98 $\pm$ 0.11 & 28.16 $\pm$ 0.06\\
GCA & TR22 + NLM180 
& 38.30 & 31.20 & 34.38 & 27.97 & 15.14 & 19.63 & 44.13 & 31.18 & 36.53 & 31.79 & 15.76 & 21.06 & 35.55 $\pm$ 0.18 & 23.32 $\pm$ 0.20 & 27.90 $\pm$ 0.17\\
\midrule
BL & {\small TR22 + NLM180 + DDI2013} 
& 29.27 & 41.93 & 34.47 & 22.93 & \textbf{25.42} & 24.11 & 38.73 & 43.79 & 41.10 & 27.11 & \textbf{27.32} & 27.21 & 29.51 $\pm$ 0.10 & 34.61 $\pm$ 0.10 & 31.72 $\pm$ 0.06\\
GCA & {\small TR22 + NLM180 + DDI2013} 
& \textbf{41.58} & 38.24 & \textbf{39.83} & \textbf{31.84} & 20.49 & 24.93 & \textbf{47.54} & 36.12 & 41.04 & \textbf{32.07} & 17.81 & 22.90 & \textbf{38.26} $\pm$ 0.16 & 28.17 $\pm$ 0.12 & 32.18 $\pm$ 0.12\\
GC$^{(2)}$ & {\small TR22 + NLM180 + DDI2013} 
& 38.85 & 36.30 & 37.52 & 29.82 & 18.59 & 22.88 & 43.74 & 34.88 & 38.80 & 31.14 & 16.40 & 21.48 & 35.89 $\pm$ 0.20 & 26.54 $\pm$ 0.20 & 30.17 $\pm$ 0.19\\
GCA + BL$^{(3)}$ & {\small TR22 + NLM180 + DDI2013} 
& 35.22 & \textbf{44.23} & 39.20 & 27.58 & 24.77 & \textbf{26.09} & 45.50 & \textbf{45.10} & \textbf{45.30} & 31.69 & 24.89 & \textbf{27.87} & 35.00 $\pm$ 0.15 & \textbf{34.75} $\pm$ 0.13 & \textbf{34.61} $\pm$ 0.10\\
   \bottomrule
   \end{tabular}}
   
   \footnotesize
   $^{(1)}$ Our original challenge submission using a BiLSTM-based approach and trained on only TR22 and NLM180.\\
   $^{(2)}$ For reference, we include an evaluation of the standard GC without attention-gating.\\
   $^{(3)}$ Our current best is a combination of GCA and BL by ensembling.
\end{table*}

\begin{table*}[!t]
  \caption{Comparison of our method with comparable (based on training data) methods of teams in the top 5.\label{tb_comparison}}
  \renewcommand{\arraystretch}{1.6}
  \resizebox{\textwidth}{!}{
  \begin{tabular}{@{\extracolsep{1em}}ll ccc ccc ccc ccc ccc}
    \toprule
     \, & \, & \multicolumn{3}{c}{Test 1 / Entity} & \multicolumn{3}{c}{Test 1 / Relation} & \multicolumn{3}{c}{Test 2 / Entity} & \multicolumn{3}{c}{Test 2 / Relation}\\
     \cline{3-5}\cline{6-8}\cline{9-11}\cline{12-14}\cline{15-17}
     \textbf{Method} & \textbf{Training Data} & \textbf{P} (\%) & \textbf{R} (\%) & \textbf{F} (\%) & \textbf{P} (\%) & \textbf{R} (\%) & \textbf{F} (\%) & \textbf{P} (\%) & \textbf{R} (\%) & \textbf{F} (\%) & \textbf{P} (\%) & \textbf{R} (\%) & \textbf{F} (\%)\\
\midrule
\citet{dandala2018ibm} & TR22 + NLM180 & 41.94 & 23.19 & 29.87 & 25.24 & 16.10 & 19.66 & 44.61 & 29.31 & 35.38 & 22.99 & 16.83 & 19.43\\
\citet{tran2018multitask} & TR22 + NLM180 & 29.50 & 37.45 & 33.00 & 22.08 & 21.13 & 21.59 & 36.68 & 40.02 & 38.28 & 22.53 & 21.13 & 23.55\\
BL + GCA (Ours) & TR22 + NLM180 & \textbf{32.89} & \textbf{41.06} & \textbf{36.51} & \textbf{24.66} & \textbf{21.35} & \textbf{22.87} & \textbf{40.57} & \textbf{42.44} & \textbf{41.47} & \textbf{28.15} & \textbf{22.42} & \textbf{24.95}\\
   \bottomrule
   \end{tabular}}
\end{table*}


In order to assess model performance with confidence intervals and draw conclusions based on statistical significance, we perform a technique called \emph{bootstrap ensembling} proposed by~\citet{kavuluru2017extracting}. That is, for each neural network (NN), we train a pool of 30 models each with a different set of randomly initialized weights and training-development set split. Performance of the NN is evaluated based on computing the 95\% confidence interval around the mean F1 of $N=100$ ensembles, where each ensemble is assembled from a set of ten models randomly sampled from the pool. This approach allows us to better assess average performance which is a nontrivial task given the high variance nature of models learned with limited training data. Our method for model ensembling (by ``voting'') is described in Section~\ref{sec-ensembling}.

We present the main results of this study in Table~\ref{tb_results} where we compare our prior efforts using strictly BiLSTMs (BL) and our current best results with graph convolutions (GCA). BL with TR22 and NLM180 as training data corresponds to our prior best at 28.16\% overall F1, while GCA with TR22, NLM180, and DDI2013 as training data represents our current best at 32.18\% overall F1 based on graph convolutions. Here, we observe a 4 point gain in overall F1 (statistically significant at 95\% confidence level based on non-overlapping confidence intervals), with most gains owing to a substantial improvement in entity recognition performance. We note that GCA is more precision focused while BL is more recall focused; moreover, GCA tends to exhibit better performance on Test Set 1, while BL tends to exhibit better performance on Test Set 2. This hints that the two architectures are highly complementary and may work well in combination. Indeed, when combined via ensembling, we observe a major performance gain across almost all measures. Here, for each ensemble, we sample five models from each pool of models (GCA and BL) for a total of ten models to ensure that results remain comparable. The resulting hybrid model exhibits the best performance overall, improving over the prior best by two points and over the current best by six points in overall F1 at 34.61\%. These differences are statistically significant at the 95\% confidence level. Next, we highlight that a main benefit of the GCA model is that it operates well with very small amounts of training data, as evident by the almost 2 absolute point improvement over the BiLSTM model when trained solely on TR22. These gains tend to be less notable when we involve examples from NLM180 and DDI2013. Lastly, we note that GCA (graph convolution with attention-gating) performs better than the standard GC (graph convolution \emph{without} attention-gating) by two absolute points in overall F1 with improvements that are consistent across all metrics. We present a comparison of our results with other works in Table~\ref{tb_comparison}. 
We omit results by \citet{tang2018two} as they are not directly comparable to ours given the stark difference in available training data. When training on strictly TR22 and NLM180 (thus being comparable to most prior work), our model exhibits state-of-the-art performance across all metrics on either test sets.

\begin{figure}[t]
\begin{minipage}{.48\textwidth}
\includegraphics[width=\linewidth]{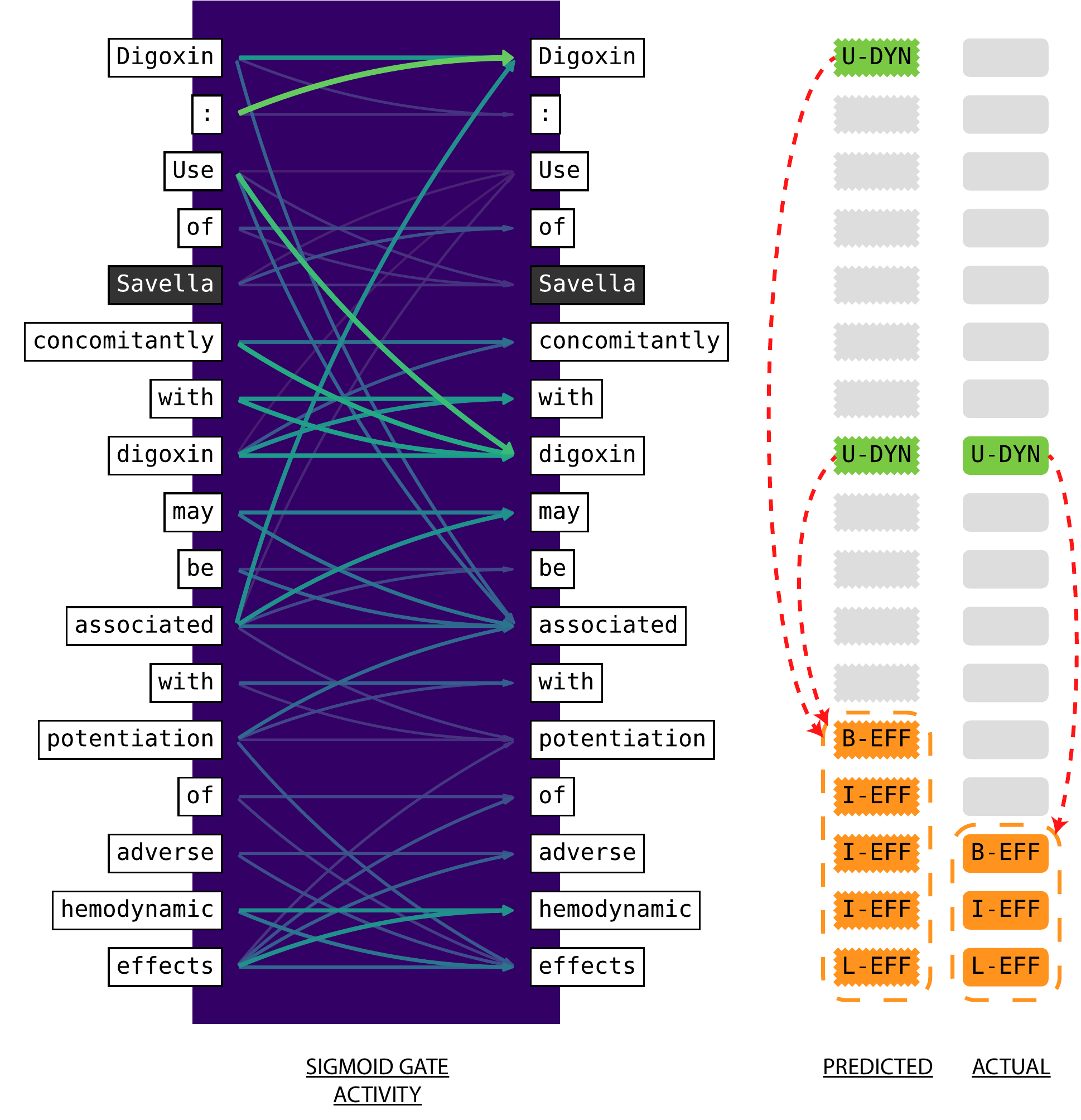}
\caption{An example sentence from the drug label for Savella along with the resulting prediction and ground truth labels. Red arrows indicate interaction outcome.}\label{fig_Savella1439}
\end{minipage}
\quad
\begin{minipage}{.48\textwidth}
\includegraphics[width=\linewidth]{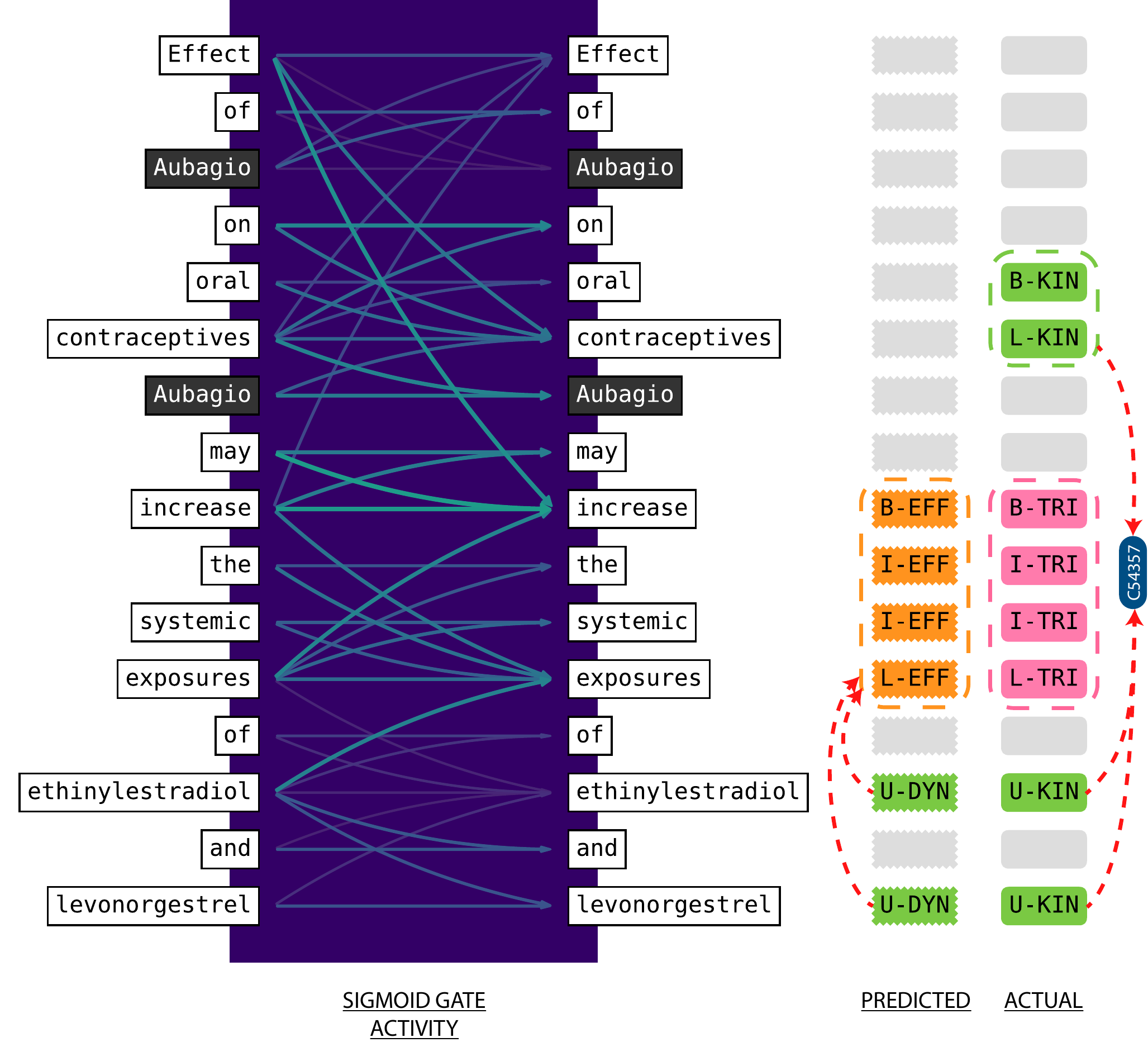}
\caption{An example sentence from the drug label for Aubagio along with the resulting prediction and ground truth labels. Red arrows indicate interaction outcome, where C54357 is a PK label corresponding to the NCI Thesaurus code for ``Increased Concomitant Drug Level.''}\label{fig_Aubagio3153}
\end{minipage}
\end{figure}

We present Figures~\ref{fig_Savella1439} and~\ref{fig_Aubagio3153} to illustrate error cases to be discussed later in Section~\ref{sec-errors}. In additional to actual and predicted annotations, these figures include a sigmoid gating activity visualization for edges in the dependency tree. The visualization serves two purposes. First, it confirms the intuition for this particular design and, second, provides a means to interpret model decisions. That is, we can observe the importance of each edge in the dependency tree as deemed by the network for a particular example. In Figure~\ref{fig_Savella1439}, for example, we can observe that for the target word ``digoxin'' (which is a precipitant, the second occurrence in the sentence), the phrase ``use'', ``concomitantly'', and ``with'' show very high activity. Likewise, signal flow from ``hemodynamic'' to ``effects'' is strong, and vice versa. Less important words such as articles appear to receive less incoming activity overall, even through self-loops. 


\section{Error Analysis}\label{sec-errors}

In this section, we perform error analysis to identify challenging cases typically resulting in erroneous predictions by the model. One major source of difficulty for the model is boundary detection in cases of multi-word entities. Errors of this type are especially prominent in case of \emph{effect} mentions which may manifest as potentially long noun phrases. Phrases with conjunctions or punctuation marks (or a combination of) may also present an obstacle for the model; for example, an \emph{effect} expressed as ``serious and/or life threatening reaction'' may instead be predicted as simply ``life threatening reaction.'' Figure~\ref{fig_Savella1439} shows a general case of this error where the model recognizes ``potentiation of adverse hemodynamic effects'' as the effect while the ground truth identifies the effect as simply ``adverse hemodynamic effects.'' This leads to both a false positive and a false negative for both the NER and the RE evaluation. We note that, given the potentially limitless ways an \emph{effect} may be expressed, any disagreement among annotators (for cases beyond those addressed in annotator guidelines) during the initial annotation process will lead to inconsistent ground truth data and thus negatively affect downstream model performance. As an example, consider the following two sentences that appear in TR22: ``Co-administration of SAMSCA with \texttt{potent CYP3A inducers} ..'' and ``For patients chronically taking potent \texttt{inducers of CYP3A}, ..'' Here, one sentence is annotated such that \emph{potent} is included as part of the precipitant expression, while another is annotated such that this modifier is excluded. 

Mixed signals and noisy labels in general tend to be an issue especially when there is limited training data as deep learning models are prone to overfitting. When evaluating on purely effect mentions, we obtain a micro-F1 score of 66\% (54\% Precision, 87\% Recall). However, the micro F1 is 87\% when ignoring the starting boundary offset and 86\% when ignoring the ending boundary offset during evaluation corresponding to roughly 20\ absolute micro-F1 gain in performance. When applying the same looser evaluation criteria to triggers and precipitants, the gains are only $\approx 6\%$ and $\approx 5\%$ respectively. Thus there is immense potential for improving entity recognition of \emph{effect} mentions if we can better handle boundary detection, possibly via rule-based methods or post-processing adjustments, with the added benefit of improving consequence prediction performance for PD interactions.  

Precipitants interacting with the label drug being mentioned multiple times may also cause issues for the model. As an example, consider the sentence presented in Figure~\ref{fig_Savella1439}. Our model identifies both mentions of the precipitant ``Digoxin'' as being involved in an interaction with the drug Savella; however, the ground truth more specifically recognizes the second mention as the sole precipitant. This results in an additional false positive with respect to both NER and RE evaluation. Lastly, there are cases where the model will mistake a mention subtly referring to the label drug as a precipitant. This is a common occurrence in cases where the label drug is not referred to by name, but by a class of drugs. Typically, identifying a mention as a reference to the drug label beforehand will disqualify it from being predicted as a precipitant. While we do use a lexicon of drug names mapped to drug synonyms and drug classes to identify these indirect mentions, it is not exhaustive for all drugs. For example, within the label of the drug Lexapro, consider the sentence ``Altered anticoagulant effects, including increased bleeding, have been reported when SSRIs and SNRIs are coadministered with warfarin.'' Here, the model recognized SSRI and SNRI as precipitants. This is incorrect, however, as Lexapro is an SSRI and these mentions are more than likely referring to Lexapro. Without this information, the model likely assumes that it is an implicit case where the label drug is not mentioned and therefore assume all drug mentions are precipitants. Hence, curating a more exhaustive lexicon for indirectly mentions of the label drug will improve overall performance.

\begin{table}[h]
\caption{Confusion matrix for interaction type\label{tb_confusion}}
\renewcommand{\arraystretch}{1.2}
\begin{tabular}{@{}rr|ccc@{}}
 & & \multicolumn{3}{c}{\underline{Predicted}}\\
 & & PD & PK & UN\\
\midrule
\multirow{3}{*}{\rotatebox{90}{\underline{Actual}}}
& PD & 788 & 37 & 68\\
& PK & 57 & 353 & 147\\
& UN & 170 & 10 & 599\\
\end{tabular}
\end{table}

Lastly, we describe a source of difficulty stemming from incorrectly classifying interaction types. Figure~\ref{fig_Aubagio3153} presents an example sentence where our model mistakes PK for PD interactions and a \emph{trigger} mention for an \emph{effect} mention. As PD and PK interactions tend to frequently co-occur with \emph{effect} and \emph{trigger} mentions respectively, predicted annotations tend to be polarized toward one pair (PD with \emph{effect}) or the other (PK with \emph{trigger}). Hence, differentiating between types of interactions for each recognized precipitant is another interesting class of error. Among all correctly recognized precipitants (based purely on boundary detection), we analyzed cases where one type of interaction, among PD, PK, and Unspecified (UN), is mistaken for another via the confusion matrix in Table~\ref{tb_confusion}. Clearly, many errors are due to cases where (1) we mistake \emph{unspecified} precipitants for PD precipitants and (2) we mistake PK precipitants for \emph{unspecified} precipitants. We conjecture that making precise implicit connections (not only whether there is evidence in the form of trigger words or phrases, but whether the evidence concerns the particular precipitant) is highly nontrivial. Likely, this aspect may be improved by inclusion of more high quality training data. Confusion between \emph{trigger} and \emph{effect} mentions is less concerning; among more than 1000 cases, there are six cases where we mistake \emph{effect} for \emph{trigger} and 20 cases where we mistake \emph{trigger} for \emph{effect}. 

\section{Conclusion}
In this study, we proposed an end-to-end method for extracting drugs and their interactions from drug labels, including interaction outcome in the case of PK and PD interactions. The method involved composing various intermediate representations including sequential and graph based context, where the latter is produced using a novel attention-gated version of the graph convolution over dependency parse trees. The so called graph convolution with attention-gating (GCA), along with transfer learning via serial pre-training using other annotated DDI datasets including DDI2013, resulted in an improvement over our original TAC challenge entry by up to 6 absolute F1 points overall. Among comparable studies (based on training data composition), our method exhibits state-of-the-art performance across all metrics and test sets. Future work will focus on curating more quality training data and leveraging semi-supervised methods overcome the scarcity in training data.

\section*{Acknowledgements and Funding}
This research was partially conducted during TT's participation in the Lister Hill National Center for Biomedical Communications (LHNCBC) Research Program in Medical Informatics for Graduate students at the U.S. National Library of Medicine, National Institutes of Health. HK is supported by the intramural research program at the U.S. National Library of Medicine. RK and TT were supported by the U.S. National Library of Medicine through grant R21LM012274.

\bibliographystyle{ACM-Reference-Format}
\bibliography{scoonerdb}

\end{document}